\documentclass[letterpaper, 10 pt, conference]{ieeeconf}  

\IEEEoverridecommandlockouts                              

\usepackage[english]{babel}
\usepackage{amsmath}
\usepackage{amssymb}
\usepackage{booktabs}
\usepackage{caption}
\usepackage{subfigure}
\usepackage[dvips]{graphicx}
\usepackage{multirow}
\usepackage{amsfonts}
\usepackage{breakurl}
\usepackage{enumerate}
\usepackage{tabularx}
\usepackage{algorithm,algorithmic}
\usepackage{bm}
\usepackage{enumerate}
\usepackage{adjustbox}
\usepackage{xcolor}
\usepackage{siunitx}
\usepackage{booktabs}
\usepackage{mdframed}
\usepackage{adjustbox}
\usepackage{authblk}
\usepackage{color}
\usepackage{xurl}
\usepackage{cite}
\makeatletter
\let\NAT@parse\undefined
\makeatother
\usepackage{hyperref}
\usepackage{relsize}
\usepackage{float}
\usepackage{pifont}

\title{\LARGE \bf
Visual Environment Assessment for Safe Autonomous \\Quadrotor Landing
}

\author{Mattia Secchiero$^{1,2}$, Nishanth Bobbili$^{1}$, Yang Zhou$^{1}$, and Giuseppe Loianno$^{1}$
\thanks{$^{1}$The authors are with the New York University, Tandon School of Engineering, Brooklyn, NY 11201, USA.
        {\tt\footnotesize email:\{ms15143, 
nb3553, yangzhou, loiannog\}@nyu.edu.}}
\thanks{$^{2}$The author is with the Department of Information Engineering, University of Padua, 35131 Padua, Italy. 
{\tt\footnotesize email:\{mattia.secchiero\}@studenti.unipd.it.}}
\thanks{This work was supported by the NSF CAREER Award 2145277, the DARPA YFA Grant D22AP00156-00, Qualcomm Research, Nokia, and NYU Wireless.}
}

\begin{document}

\maketitle
\thispagestyle{empty}
\pagestyle{empty}

\begin{abstract}
Autonomous identification and evaluation of safe landing zones are of paramount importance for ensuring the safety and effectiveness of aerial robots in the event of system failures, low battery, or the successful completion of specific tasks.
In this paper, we present a novel approach for detection and assessment of potential landing sites for safe quadrotor landing. Our solution efficiently integrates 2D and 3D environmental information, eliminating the need for external aids such as GPS and computationally intensive elevation maps. The proposed pipeline combines semantic data derived from a Neural Network (NN), to extract environmental features, with geometric data obtained from a disparity map, to extract critical geometric attributes such as slope, flatness, and roughness. We define several cost metrics based on these attributes to evaluate safety, stability, and suitability of regions in the environments and identify the most suitable landing area.
Our approach runs in real-time on quadrotors equipped with limited computational capabilities. Experimental results conducted in diverse environments demonstrate that the proposed method can effectively assess and identify suitable landing areas, enabling the safe and autonomous landing of a quadrotor.

\end{abstract}



\section*{Supplementary Material}
\noindent \textbf{Video}: \url{https://youtu.be/3tH621vF8LM}
\section{Introduction}\label{sec_I}


Unmanned Aerial Vehicles (UAVs) have become increasingly popular platforms to assist humans in several complex and dangerous applications such as surveillance, law enforcement, mapping, search and rescue, delivery services and precision agriculture~\cite{LAGHARI20238, https://doi.org/10.1002/net.21818}.
The development of novel autonomous algorithms coupled with the drop in price–performance ratio of processors and sensors supported also the execution of complex tasks such as collaborative transportation~\cite{li2023safetyaware}, autonomous flight~\cite{8279692}, collision avoidance~\cite{7105819}, exploration~\cite{9721080} as well as shipping and delivery~\cite{TANG20191} or industrial inspection~\cite{drones7080515}.
To ensure the safety of individuals, structures, and overall mission success in the aforementioned applications it is commonplace to equip aerial robots with intelligent landing capabilities~\cite{10.1145/3447587.3447590, schoppmann2021multiresolution, article}. These mitigate the risks posed by mechanical and sensory failures, ensuring secure operations in challenging scenarios. This not only minimizes potential threats posed to individuals and structures, but also enables the successful execution of tasks that require such capabilities. For example, in the case of a drone delivery system that operates in urban environments, the ability to accurately identify suitable landing zones becomes crucial for the successful delivery of packages. In agriculture, drones play a pivotal role in monitoring crops, assessing plant health, and optimizing agricultural practices. Upon completing these tasks, the drone requires a reliable and safe landing procedure.
 \begin{figure}
     \centering
     \vspace{0.7em}
     \includegraphics[width=1\linewidth]{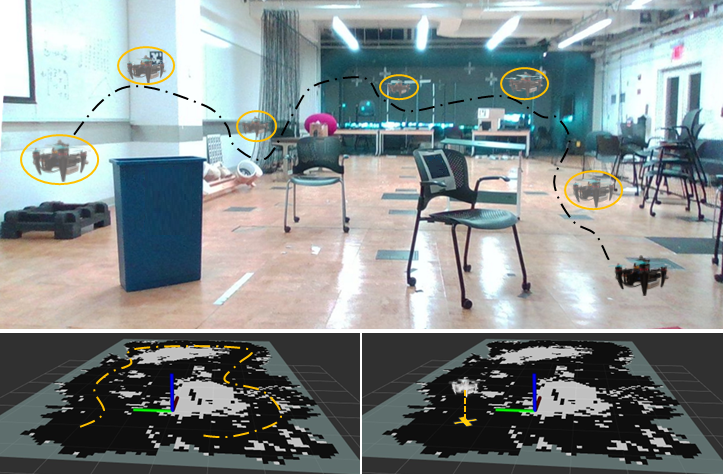}\\

     \caption{
    \textit{Top}: our drone navigating and mapping the environment. \textit{Bottom}: the associated 2D binary map of the safe and unsafe landing locations (\textit{left}) and the chosen safe landing spot in the 2D map (\textit{right}). The black areas are the safe landing locations, while the gray ones are unsafe.
     }
     \label{fig:overview}
     \vspace{-20pt}
\end{figure}

However, current state-of-the-art solutions tend to be fragile and computationally intensive, often require preliminary environment knowledge and still offer limited autonomy.
Similarly, the majority of commercially available landing solutions often relies on manually designed, pre-defined navigation policies to aid humans in guiding the robots to land near the intended or required location, therefore offering minimal or no autonomy also for the landing process.

This paper presents several significant contributions.
First, we propose a novel visual environment detection and assessment approach for the safe autonomous landing of aerial robots. Compared to the state-of-the-art solutions,  our method efficiently combines metric and semantic information, leveraging both RGB images and the disparity maps extracted from the environment.
Specifically, the 3D points of the environment, generated from the disparity map, are projected onto the segmented RGB image to efficiently process only the ones associated with safe regions, obtaining an effective and efficient solution.
Furthermore, our method does not need to build and store computationally intensive elevation maps. Instead it directly generates and updates a 2D binary map of safe and unsafe landing zones, without sacrificing any relevant information compared to existing approaches.
Second, we define several cost metrics based on critical 
geometric attributes such as slope, flatness, and roughness, extracted from the visual information to assess potential viable landing areas. 
Finally, our pipeline operates on-board, without relying on any off-board streaming of data, GPS or pre-obtained information. We demonstrate that our framework successfully enables safe quadrotor landings in multiple and challenging indoor environments.


\section{Related Works}
Several works address the problem of identifying a safe landing zone and implementing autonomous landing procedures. The vast majority of the existing approaches are vision-based, given the Size, Weight, and Power (SWaP) constraints of small-size aerial robots. For example in~\cite{1570727} the authors compute the local slope and roughness of a Digital Elevation Model (DEM) to detect and avoid hazards such as steep slopes, rocks, cliffs, and gullies. Another approach, as described in~\cite{inproceedings1}, defines a cost function that evaluates the physical properties of the local neighborhood within an elevation map region to identify safe landing areas, using a robot-centric fixed size map. This, however, confines the environmental knowledge exclusively to the region beneath the drone.
The solution proposed in~\cite{schoppmann2021multiresolution} instead infers a safe landing zone by evaluating slope and roughness from the DEM of the environment, obtained using a Structure from Motion (SfM) algorithm.
Conversely, the authors in~\cite{10.1145/3447587.3447590} apply a slope and roughness threshold to the DEM image gradient, to only retain flat terrains. Since the elevation maps are pre-determined, a Neural Network (NN) segments the RGB images of the possible landing area as a final evaluation step. The semantic information is used to asses the validity of the area and to overcome the fact that the maps could be outdated.
Other works such as~\cite{article} and~\cite{tomita2022bayesian} directly implement semantic segmentation on a DEM. Segmentation is employed to classify hazardous and safe landing locations without resorting to plane fitting techniques or gradient thresholding. However, these approaches rely on pre-collected LiDAR data to get the elevation models, which may not be suitable for small-size UAVs and might not always have access to up-to-date maps.


Yet, it is essential to develop general-purpose solutions that do not rely on GPS~\cite{1570727, 10.1145/3447587.3447590} or pre-determined maps~\cite{10.1145/3447587.3447590, article, tomita2022bayesian}. The former would be unsuitable for indoor or GPS-denied environments, while the latter's reliance on pre-defined environment poses a challenge in dynamic settings, potentially leading to catastrophic outcomes.~\cite{inproceedings1} and~\cite{schoppmann2021multiresolution} further illustrate this by employing a fixed-size map, limiting environmental awareness to the area directly below the drone, thus disregarding a big portion of the overflown area. In~\cite{7008097}, the 3D information are lacking, consequently failing to comprehensively address crucial factors like slope, flatness, and roughness.

Compared to the aforementioned existing solutions, we directly construct a variable dimension 2D binary map to guide the drone toward a safe landing location. In such a way, our approach does not need to derive any elevation map, resulting in a lighter and more efficient implementation, while still evaluating all the relevant aspects related to the safe site detection. In addition, our pipeline stands out by its independence from external aids such as GPS, off-board or pre-obtained geometric information and autonomously implements inspection and landing behaviours.
\section{Methodology}

 \begin{figure*}
     \centering
     \includegraphics[width=1\linewidth]{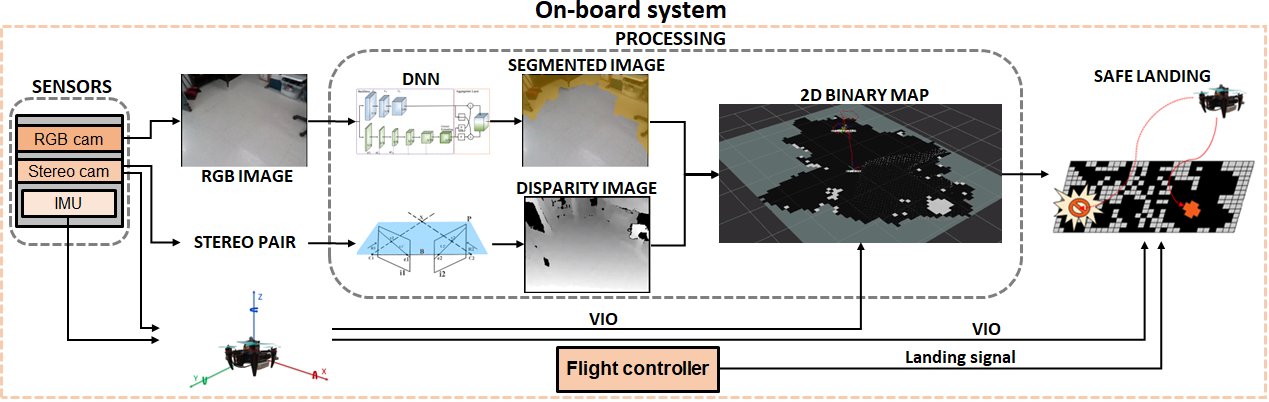}\\
     \caption{
    Overview of our autonomous safe site detection and landing system: we use our quadrotor with a NVIDIA Jetson NX for computation and a stereo camera for VIO \& mapping the environment. All our algorithms run in real-time onboard.
     }
     \label{fig:pipeline_safe}
     \vspace{-20pt}
\end{figure*}

 
Figure~\ref{fig:pipeline_safe} gives an overview of the entire system's pipeline.
The drone leverages a stereo camera pair combined with data coming from an Inertial Measurement Unit (IMU) to compute Visual-Inertial Odometry (VIO). The real-time estimation of the drone's position and orientation, with respect to the fixed world reference frame ($\mathcal{F}_{W}$), is essential for autonomous flight capabilities such as exploration and autonomous landing.
The 2D occupancy grid, on the other hand, is generated considering the segmented RGB images and the disparity maps obtained from a stereo pair. This occupancy grid directly embeds both semantic and geometric information, representing the safe and unsafe landing regions of the environment. 
At the perception level, one key distinction between our approach and other methodologies lies in the way we generate the map of safe and unsafe landing locations. In most other approaches, the workflow involves the initial construction of a DEM or an elevation map. Subsequently, a binary map of the environment is created, and safe landing areas are identified within this map.
In contrast, our method simplifies this process by directly creating a $2$D variable-dimension occupancy grid, without the need for elevation maps. This grid encodes a binary classification, distinguishing safe and unsafe landing locations, while still retaining the essential $3$D information. Furthermore, it also employs both metric and semantic information, unlike many other approaches that rely solely on one type of information, increasing the robustness of the overall solution.
The grid is dynamically updated based solely on the safe point cloud data that meet all the safe site criterion. 

Finally, based on this representation, in the evaluation step we find the actual landing zone. From the $2$D map and the drone's position we locate the best landing zone by minimizing a cost function that considers \textit{(a)} the drone's distance from a potential safe area and \textit{(b)} the distance of the closest unsafe point to a possible safe area.
This process is iterated across the entire map to find a safe zone, large enough to accommodate the drone during landing while also ensuring a safety margin.
Once the inspection behaviour concludes, or the necessity of landing arises, the safe landing coordinates are retrieved and the landing behaviour is performed autonomously.



\subsection{Safe Site Detection}
\label{subsec:safesite}

\subsubsection{\textbf{Semantic Information}}
The RGB images are segmented through BiSeNetV2~\cite{yu2020bisenet}, a real-time semantic segmentation neural network. Compared to an encoder-decoder structure or the pyramid pooling modules often used in semantic segmentation, this network proposes a bilateral structure, namely treats the spatial details and categorical semantics separately to achieve high accuracy and high efficiency for real-time segmentation tasks. The features extracted by the two branches of the dual-pathway backbone are then merged together by an aggregation layer.
Additionally, to enhance the inference time of the network, we leverage the trained BiSeNetV2 model and we optimize it through the NVIDIA TensorRT library~\cite{nvidia-tensorrt-docs}. This significantly accelerates the network's performance and also reduces the model dimensions.
The goal of the network is to recognize image regions that are suitable for landing such as grass fields, pavements, roads, floors, etc. For each pixel in the image the NN associates a semantic meaning to it
\begin{equation}
    f : (u, v) \mapsto C
\end{equation}
where at each location $(u, v)$ the pixel is characterized by a label $C$. Knowing the relation between the label $C \in \mathbb{R}^+$ and the corresponding class (e.g., $C : 0 \mapsto$ safe landing, $C : 1 \mapsto$ people, $C : 2 \mapsto$ obstacles, ...) we are able to infer if the region is suitable for landing.

\subsubsection{\textbf{Geometric Information}}
From the stereo pair instead we derive a disparity map. Utilizing the camera intrinsic and extrinsic parameters, we calculate the Cartesian coordinates ($x$, $y$, and $z$) of each point, generating the associated point cloud.
Subsequently, we re-project the point cloud on the segmented image to only keep the points associated to a safe landing region as
\begin{equation}
    \begin{bmatrix}
        u' \\
        v' \\
        w  \\
        *
    \end{bmatrix}
        = 
    \begin{bmatrix}
        f_x & 0 & c_x & 0\\
        0 & f_y & c_y & 0\\
        0 & 0 & 1 & 0\\
        0 & 0 & 0 & 1
    \end{bmatrix}
    \begin{bmatrix}
       \mathbf{R}_{SL}^{RGB} & \mathbf{t}_{SL}^{RGB} \\
       \mathbf{0} & 1
    \end{bmatrix}
    \begin{bmatrix}
           x \\
           y \\
           z \\
           1
   \end{bmatrix},
\label{eq:reproject}
\end{equation}
where $f_{x}$ and $f_{y}$ represent the focal length and $c_{x}$ and $c_{y}$ are the optical center coordinates. Starting from a point ($x$, $y$, and $z$) we determine its corresponding pixel coordinate ($u$, $v$), using the transformation equations $u = \frac{u'}{w}$ and $v = \frac{v'}{w}$. It is important to note that in our case, the RGB camera frame ($\mathcal{F}_{RGB}$) and the left stereo camera frame ($\mathcal{F}_{SL}$) are not perfectly aligned. Thus, we account for the rigid transformation between the two frames ($\mathcal{\textbf{H}}_{SL}^{RGB} \in \text{SE}(3)$), as indicated in the second term of the right-hand side of eq. \eqref{eq:reproject}. This adjustment ensures the accurate projection of 3D points onto the 2D image.
Once we have the pixel coordinates ($u$, $v$), we check if the corresponding image location falls within a safe landing region. In the positive case, the $3$D point is retained; otherwise, it is flagged as unsafe and discarded.
By only considering the safe points we speed-up the computation, improving the performance of the algorithm.

Subsequently, the point cloud is filtered and down-sampled to enhance its quality and suitability for safe landing site detection. 
The point cloud is initially down-sampled using a voxel grid filter. The "leaf size" parameter controls the voxel size, with larger values resulting in greater down-sampling. In our case, we choose to retain one point each $0.1\times 0.1~\si{m}^2$, striking a balance between processing speed and accuracy.
Then, a statistical outlier removal filter is applied in order to remove the points that significantly deviate from the heuristic distribution of the point cloud.
To further improve the point cloud quality, we employ a Moving Least Square (MLS) smoothing filter, resulting in a smoother point cloud less affected by noise.
Finally, a plane fitting algorithm ~\cite{10.1145/358669.358692} is executed to identify planar regions within the point cloud. This allows to retain only the points associated to a flat surface, that also satisfy predefined slope and roughness thresholds. 
However, to correctly assess the metric properties of the scene, we first have to ensure that the point cloud is aligned with the world reference frame ($\mathcal{F}_{W}$), which conveniently coincides with our map reference frame ($\mathcal{F}_{2DM}$). To this end we consider the rigid transformation between the left stereo frame and the world frame $\mathcal{\textbf{H}}_{SL}^{W} \in SE(3)$, since the point cloud is originally aligned with $\mathcal{F}_{SL}$. By leveraging the drone's odometry, we can compute $\mathcal{\textbf{H}}_{SL}^{W}$ and apply the necessary compensation to get the true plane's inclination. Moreover, in this way we also compensate for the drone's Roll, Pitch, and Yaw (RPY) rotation since it can be tilted with respect to the surface.
Subsequently, the plane inclination can be computed considering the angle between its normal vector $\hat{\mathbf{n}} = (\mathbf{n_x}, \mathbf{n_y}, \mathbf{n_z})$ and the $\mathbf{z}$-axis of the world frame (parallel to the gravity vector $\hat{\mathbf{g}}$) as
\begin{equation}
    \phi_x = \operatorname{atan2}(\mathbf{z}, \mathbf{n_y}),~\phi_y = \operatorname{atan2}(\mathbf{z}, \mathbf{n_x}).
\end{equation}
In our specific setup, any planes with inclinations exceeding $15$° are considered unsafe.
Furthermore, we can establish whether a point qualifies as a plane inlier or not by evaluating its distance to the fitted plane, thus defining the maximum roughness admitted.
To retrieve the distance of a point from the fitted plane, we compute the magnitude of the perpendicular vector connecting the point to the plane as 
\begin{equation}
    \text{dist} = \frac{|Ax + By + Cz + D|}{\sqrt{A^2 + B^2 + C^2}},
\end{equation}
where $x, y$ and $z$ are the point's coordinates and $A, B, C$ and $D$ are the plane coefficients.
In our case if a point is more than $0.05$ m away from the plane it is considered unsafe.
The map is a composition of processed images and point clouds that satisfy the safe site criteria detailed in this section.

\subsection{Environment Assessment and Autonomous Landing}
\label{subsec:finaleval}


\subsubsection{\textbf{Environment Assessment}}
Up to this point, our knowledge has provided us with a general understanding of the distribution of safe and unsafe points across the map. However, our objective is to pinpoint the optimal landing location.
To achieve this, we search for a patch on the map that spans an area approximately $1.85$ times the size of the drone. This patch must be a sufficiently large region consisting exclusively of safe points and accounting for an additional safety margin.
The patch is considered safe when every cell within it has a safety probability "$\mathit{p}$" of $95\%$ or higher. This means that we ensure all cells in the designated patch are highly likely to be safe, with safety probabilities ranging from $0\%$ to $100\%$, where $100\%$ indicates complete safety.
The probability "$\mathit{p}$" of each cell is computed as done in~\cite{hornung13auro}.
Once this area is found, we compute the associated cost
\begin{equation}
     J = \mathbf{\alpha} \cdot J_{d} + \mathbf{\beta} \cdot \frac{1}{J_{un}},
     \label{cost_fcn}
\end{equation}
with $\mathbf{\alpha} + \mathbf{\beta} = 1$ and $\mathbf{\alpha},~\mathbf{\beta} \in [0,1]$, $J_{d}$ is the distance between the center of the safe zone and the drone while $J_{un}$ is the distance between the center of the safe zone and the closest unsafe point.
$J_{d}$ and $J_{un}$ are both computed as Euclidean distances.

This approach prioritize landing zones that are not only closer to the drone, but also farther away from unsafe areas. Fine-tuning the parameters $\alpha$ and $\beta$ enables us to adjust the behavior for identifying the safest landing zone. By increasing $\alpha$ and reducing $\beta$, the algorithm tends to find a safe landing zone that is closer to the drone, but potentially nearer to obstacles, and vice versa.
Additionally, our final evaluation takes into account the drone's battery consumption, which is influenced by the Euclidean distance between the drone and the 3D landing location, as shown in~\cite{6884813}.
By iterating this last step over the whole map we aim to minimize the cost function and only keep the safest landing zone, namely the one associated to the lowest cost possible.
Whenever new areas are overflown or the drone changes its position, the environment is re-perceived, the map is updated accordingly and the best safe landing zone is published.

In summary, our safe site detection pipeline employs a comprehensive evaluation of region safety, taking into account both semantic information and geometric information, including flatness, roughness, steepness, a distance transform and the drone size.
\subsubsection{\textbf{Autonomous Landing}}
The landing step is pretty straightforward and doesn't require any particular process to be involved. 
Once the landing is required, a minimum snap trajectory generation algorithm~\cite{LoiannoRAL2017,MaoTRO2023} is used to find a path from the drone's actual position to the safe landing spot.
When initiated, the drone follows this two-step behaviour:
\begin{itemize}
  \item Fly above the safe landing zone, utilizing the minimum snap trajectory.
  \item Decrease the height until it reaches the ground level.
\end{itemize}
This approach is employed since no collision avoidance behaviour has been utilized, thus prioritizing safety and avoiding any possible crash.
\section{Results}
\label{cap:res}


To validate the proposed approach, we execute a series of real experiments in a challenging, large indoor environment measuring $26\times10\times4~\si{m^3}$, situated at the Agile Robotics and Perception Lab (ARPL, New York University).
In particular, our validation process involved two key aspects:
\textit{(a)} \textbf{environmental changes}, to simulate different evaluation and landing scenarios;~\noindent \textit{(b)} \textbf{waypoints diversification}, to simulate different inspection strategies.

\subsection{System Setup}
Our drone is a compact and versatile system, with a diameter of $0.27~\si{m}$ and a weight of $1.1$ kg. It is equipped with a PX4 Autopilot flight controller, for high level position control, and a Nvidia Jetson NX computing board.
For convenience, without loss of generality of our approach, we employ two stereo cameras to decouple the localization and safe landing evaluation due to the camera characteristics, introducing as well some redundancy in the system. The first one, a RealSense T265 tracking camera, employed to obtain a robust VIO and the second one, responsible for mapping the environment and detecting the safe landing zone. Depending on mission requirements and constraints, it is also possible to effectively operate with a single stereo camera, either pointing directly downward or tilted at a $45$-degrees angle, not affecting the approach and results of this work.
The system relies on the ROS middleware~\cite{inproceedings2}, facilitating communication and integration among the various modules. Considering the real-time and resource-constrained applications, these modules are managed by a nodelet, that reduces computation and latency by sharing memory space and avoiding inter-process communication overhead.

\subsection{Neural Network Training and Evaluation}
To train our network, we leverage the ADE20K semantic scene parsing dataset~\cite{zhou2018semantic}, since it provides a wide set of indoor and outdoor environments, covering a wide range of classes and examples relevant for addressing the safe landing task. 
However, rather than using the original $150$ classes, we have manually clustered them into $11$: \textit{water, people/animals, sky, trees, man-made obstacles, nature obstacles, safe landing site, light, vehicles, background, buildings}. In such a way we enhance the inference time simplifying the problem and allowing the
segmentor to generalize better, as done in~\cite{10.1145/3447587.3447590}. 
Our training pipeline is based on the PyTorch framework developed by openMMLab \cite{mmseg2020}. It consists in an iteration-based training process using a Stochastic Gradient Descent (SGD) optimizer for $160$ K iterations, and two NVIDIA GPUs with batch size equal to $8$. The optimizer parameters are reported in Table~\ref{tab:opt_param}.

\begin{table}[htbp]
\centering
\begin{tabular}{|c|c|}
\hline
\textbf{Optimizer parameters} & \textbf{Value} \\
\hline
Learning Rate & 0.05 \\
\hline
Momentum & 0.9 \\
\hline
Weight Decay & 0.0005 \\
\cline{1-2} 
Decay Type & Polynomial Decay \\
\hline
\textbf{Polynomial Decay parameters} & \textbf{Value} \\
\hline
Learning Rate$_{\min}$ & $1 \times 10^{-4}$ \\
\hline
Power & 0.9 \\
\hline
\end{tabular}
\caption{NN Training parameters}
\label{tab:opt_param}
\vspace{-10pt}
\end{table}

To improve the quality of our segmentation results, we performed a fine-tunig on the NN using a custom indoor dataset. The dataset includes approximately $1.2$ K images of common scenes within our lab environment. Notably, the environments created for testing differ from the ones used during the fine-tuning phase, since they incorporate new scenes and objects for evaluation.
Since manual labeling is a very time-consuming activity, the Segment Anything Model (SAM)~\cite{kirillov2023segment} is used to facilitate the mask creation process. SAM is designed to address the challenges of creating high-quality masks for various objects in images. It is trained on $11$M images with over $1$B masks and can produce valid segmentation masks in real-time, when prompted with different types of inputs such as points, boxes, and text. Once the masks are retrieved, we can assign the correct labels to each one of them, thus identifying the ground truth of each image.
Furthermore, during training we employ a data augmentation pipeline to increase the dataset size. This pipeline is based on random resizing, random cropping, random flipping and photometric distortion. 
The fine-tuning parameters coincide with the one specified in Table \ref{tab:opt_param}, with the exception that training continued for another $80$ K iterations. This resumed from $LR = 10^{-4}$ and finished with $LR_{\min} = 1 \times 10^{-5}$.

\begin{figure}
  \centering
  \subfigure{\includegraphics[width=0.49\linewidth]{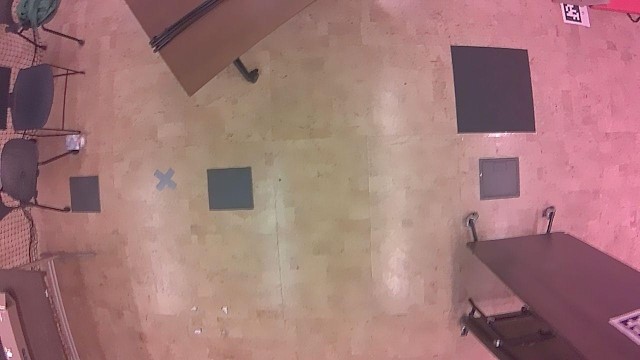}}\hfill
  \subfigure{\includegraphics[width=0.49\linewidth]{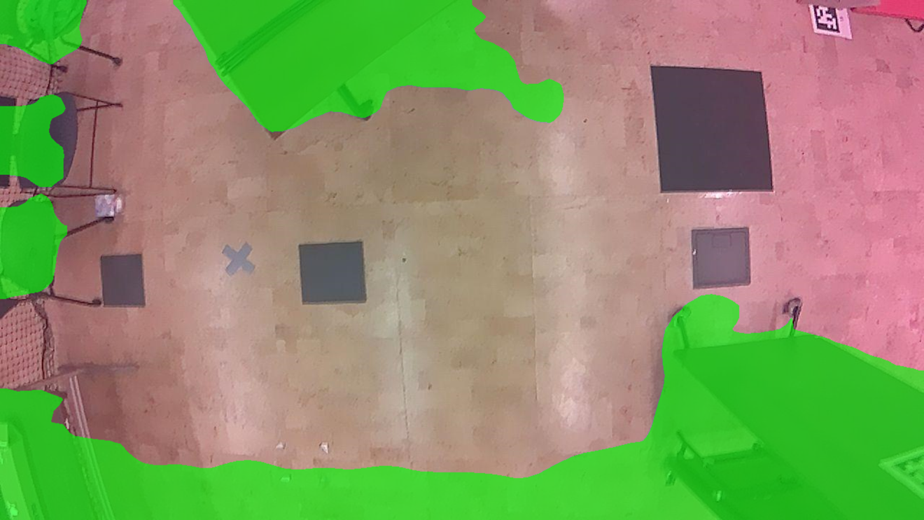}} \\
  \vspace{-5pt}
  \subfigure{\includegraphics[width=0.49\linewidth]{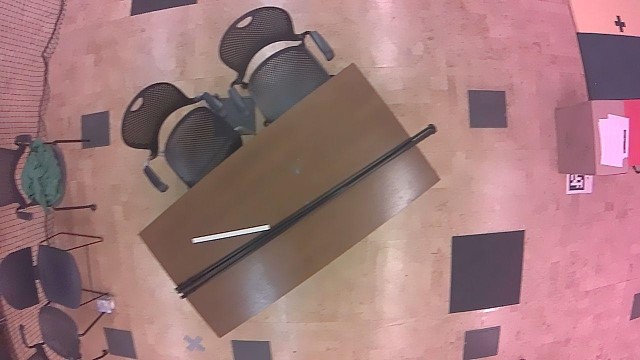}}\hfill
  \subfigure{\includegraphics[width=0.49\linewidth]{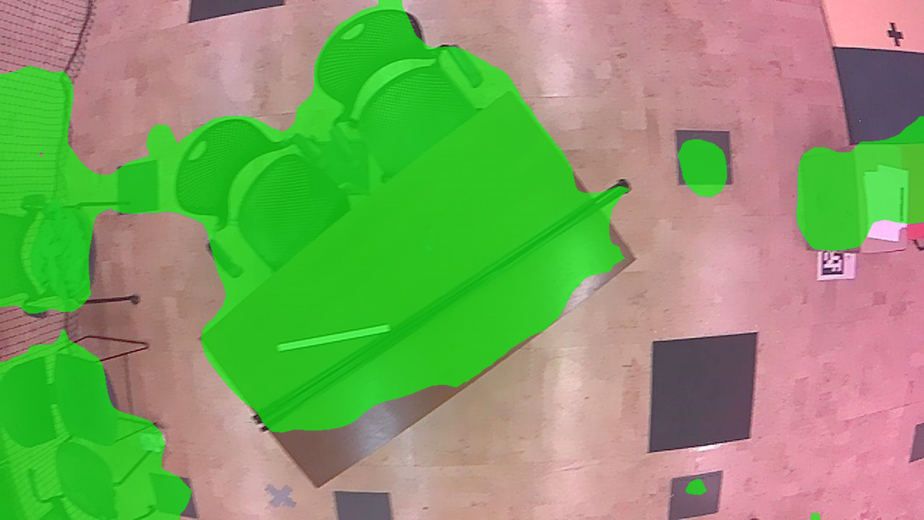}} \\
  \vspace{-5pt}
  \subfigure{\includegraphics[width=0.49\linewidth]{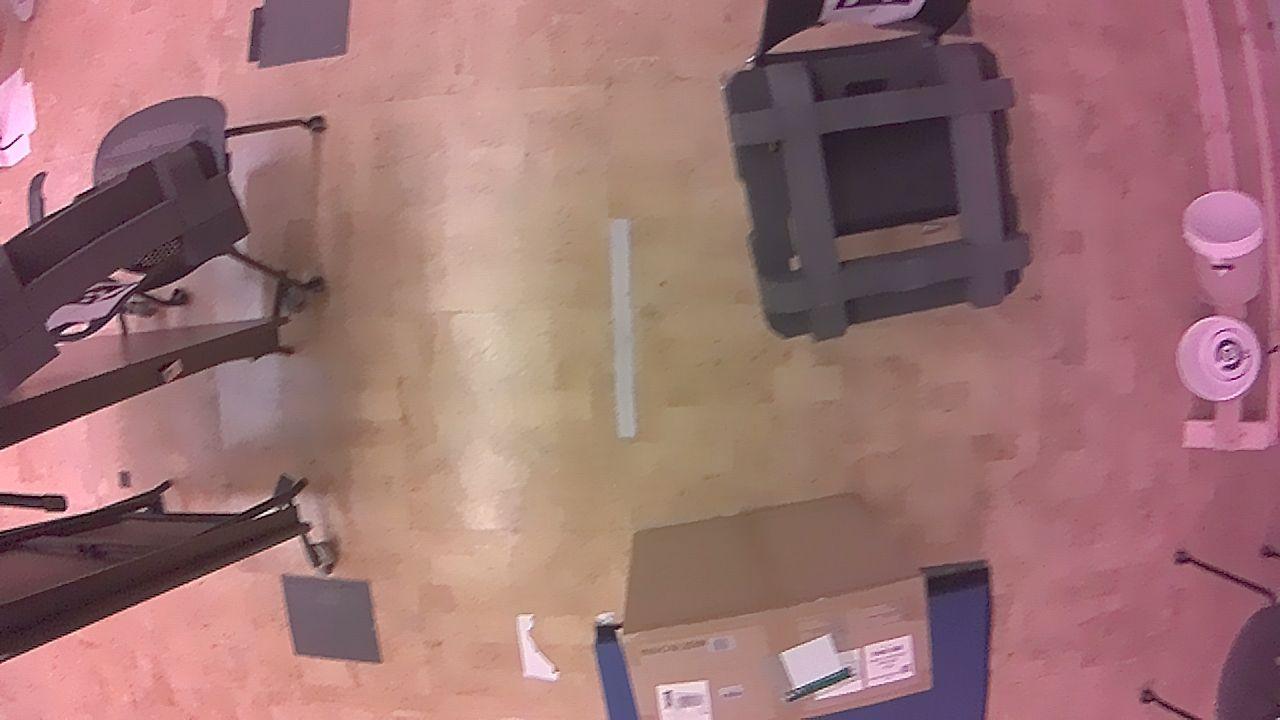}}\hfill
  \subfigure{\includegraphics[width=0.49\linewidth]{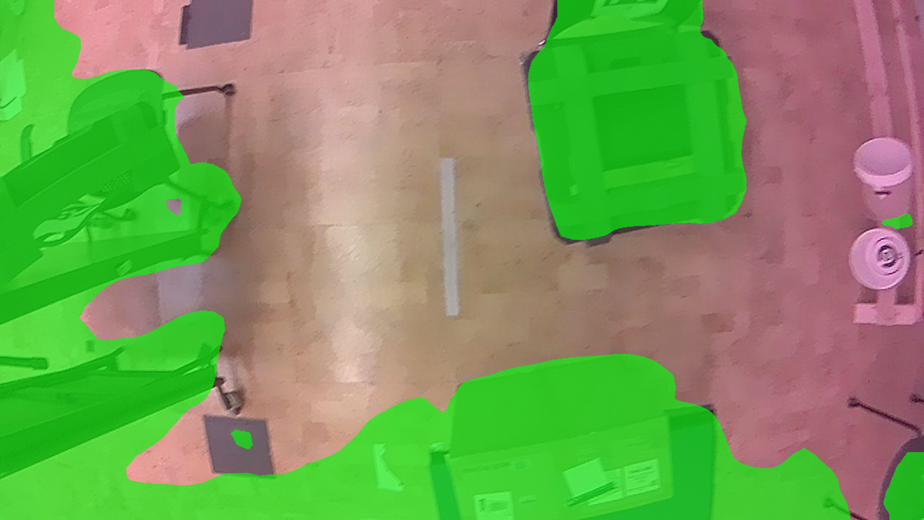}} \\
  \vspace{-5pt}
  \caption{Segmentation results in three different scenarios: on the \textit{left} column the RGB images, on the \textit{right} column the segmentation results. The green areas are considered unsafe.}
  \vspace{-20pt}
  \label{fig:images_seg_result}
\end{figure}

The NN runs on-board the Jetson NX at $7.1$ Hz and obtains a mean Intersection over Union (mIoU) of $67.51\%$ and a mean Accuracy (mAcc) of $85.21\%$. For a qualitative evaluation of the segmentation results, please refer to Fig. \ref{fig:images_seg_result}.

\subsection{Environment Assessment and Autonomous Landing}

In our test scenario, we operate with a map resolution of $0.1$ meters, while the designated safe landing zone measures $0.5\times 0.5~\si{m}^2$.
The acquisition of RGB images and the stereo pair occurs at a rate of $30$ Hz, whereas the disparity maps runs at $3$ Hz, which suffices for our operational speeds. The processed point clouds and the safe landing locations are instead published at a frequency of $1.1$ Hz. However, if better performances are required, we have the flexibility to increase the updating frequency of the processed points. Finally, VIO runs at $100$ Hz. 

For our tests, we select $\alpha=0.65$ and $\beta=0.35$. As detailed in eq. \eqref{cost_fcn}, we empirically observe that these settings prioritize the term related to the drone's proximity to the safe landing area over the distance between the safe landing site and obstacles. Moreover, the slope and roughness thresholds are set respectively to $15$° and $0.05$ m.

 \begin{figure*}
     \centering
     \includegraphics[width=1\linewidth]{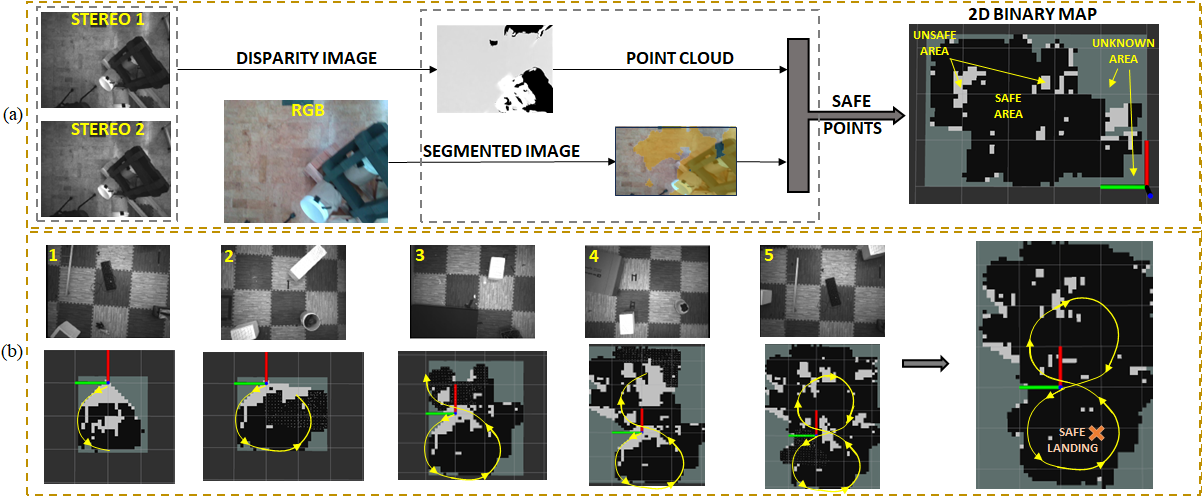}\\
     \caption{\textit{(a)} Data acquisition \& processing pipeline for the map creation and \textit{(b)} Site evaluation and safe autonomous landing experiment in a low height, middle density environment scenario with an "8" navigation pattern. }
     \label{fig:mapping_qualitative_res}
     \vspace{-15pt}
\end{figure*}
\begin{figure}
  \centering
  \subfigure[High density, low height obstacles scenario. Acc = $76.5\%$.]{\includegraphics[width=0.48\linewidth]{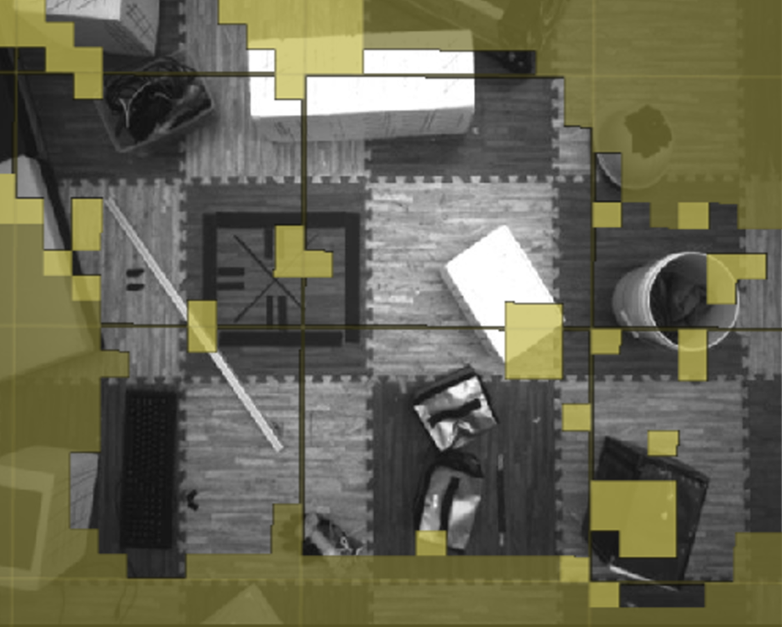}}\hfill
  \subfigure[Long waypoints navigation pattern scenario. Acc = $82.6\%$.]{\includegraphics[width=0.48\linewidth]{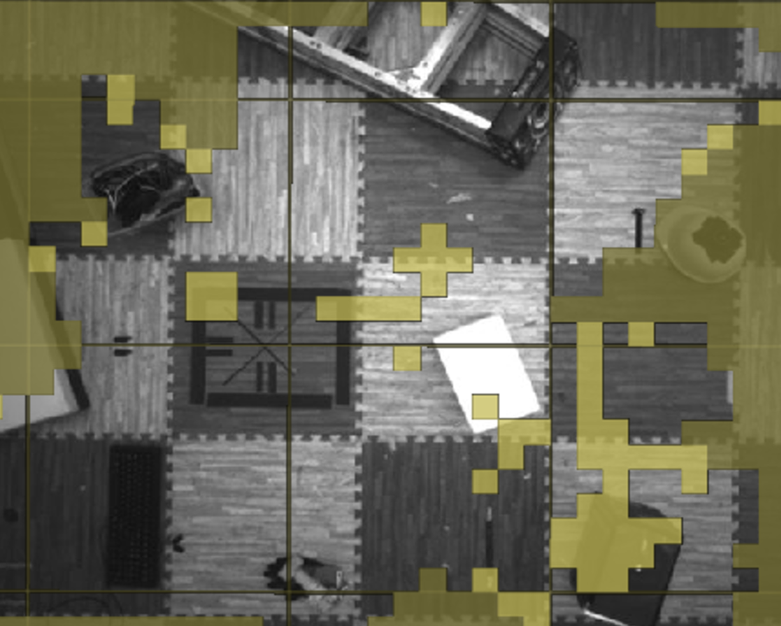}} \\
  \vspace{-5pt}
  \subfigure[Medium density, medium height obstacle scenario 1. Acc = $91.4\%$.]{\includegraphics[width=0.48\linewidth]{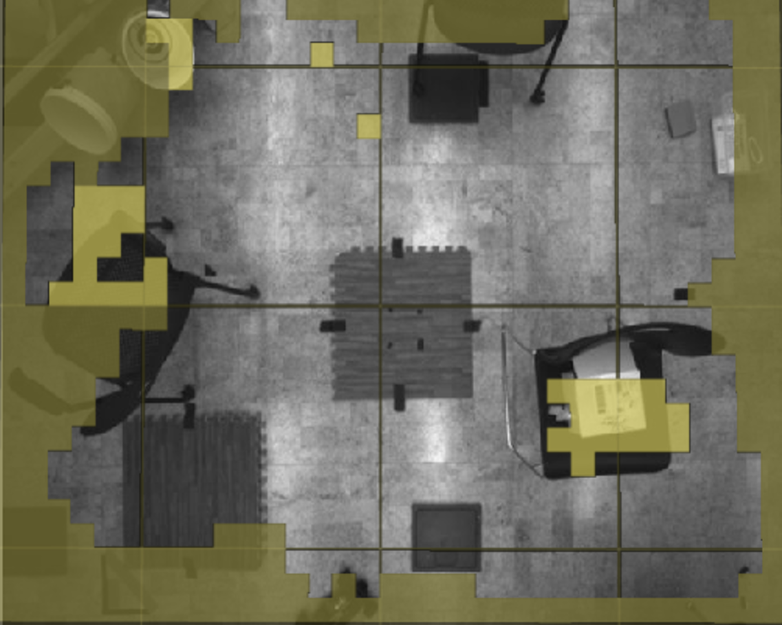}}\hfill
  \subfigure[Medium density, medium height obstacle scenario 2. Acc = $75.1\%$.]{\includegraphics[width=0.48\linewidth]{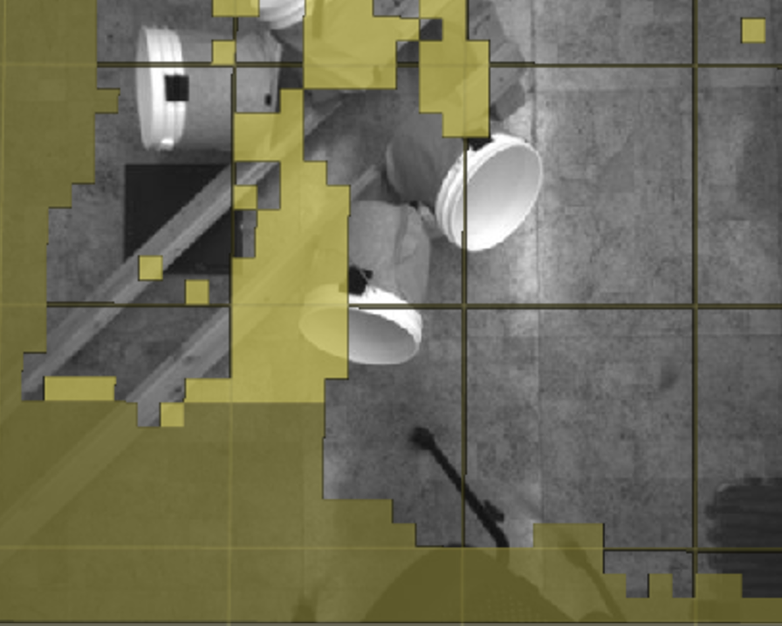}} \\
  \vspace{-5pt}
  \caption{2D binary map of safe and unsafe landing locations, overlayed to the real environment. The light green regions are unsafe while the dark green are unknown and still to be explored. Both of them are hazardous areas for landing.}
  \vspace{-20pt}
  \label{fig:quant_res}
\end{figure}

The proposed pipeline is tested in several scenarios, including different obstacle heights and densities and different navigation patterns and speeds.
In Fig. \ref{fig:mapping_qualitative_res}, we showcase the mapping process and a full experiment in a low height, middle density obstacles environment. After take-off, the drone  follows an "8" navigation pattern and performs a couple of flight runs over the environment. As we can see from the succession of images in Fig.~\ref{fig:mapping_qualitative_res}(b), each time the drone perceives new areas, by following its path, the map is updated accordingly. When the navigation behaviour is concluded, the $2$D occupancy grid is fully updated and the drone can implement the final environmental assessment. Once the safest landing area is identified, the drone finally implements the autonomous landing.
The last picture in Fig.~\ref{fig:mapping_qualitative_res}(b) shows the whole map with the safe landing zone location and the drone's path.
Out of $7$ tests performed in different challenging environments, the drone is able to safely land $6$ times, showing its capability to detect a safe landing zone with an overall success rate of $85.71\%$. The unsuccessful landing is not attributed to any errors in segmentation or metric data but rather to the grid discretization. In this specific experiment, the grid size was excessively large in comparison to certain low-height obstacles. Through testing with a slightly smaller grid, we can effectively address this issue without impacting computational efficiency. By overlaying the created $2$D occupancy grid with the top view of the environment, we can also quantitatively evaluate the number of zones correctly "classified" as safe or unsafe.
We evaluate the classification accuracy as
\begin{equation}
\text{Acc} = \frac{TP + TN}{TP + FP + FN + TN},
\end{equation}
where $TP$ are the true positive, $TN$ the true negative, $FP$ the false positive and $FN$ the false negative.
In Fig. \ref{fig:quant_res}, we show some of the results obtained during testing in four environments, considering different obstacles heights, densities, and navigation patterns.
The mean accuracy (mAcc) in identifying safe landing zones is approximately $81.4\%$.

\section{Conclusion}
In this work, we have presented a visual approach to autonomously detect safe landing sites on-board a quadrotor, with limited SWaP resources. The proposed approach allows accurate and efficient safe landing detection since it combines both semantic and metric information and directly computes a 2D binary map of the overflown environment, thus avoiding the creation of expensive elevation maps. 
Furthermore, we have also shown its ability to guarantee real-time, safe autonomous landing in real-world environments.

In the future, we are considering to 
enhance the  NN performances by adopting more sophisticated architectures, such as DeeplabV3+~\cite{chen2018encoderdecoder}. In such a way we could improve the accuracy in identifying a safe landing zone, since this will in part be limited by the maximum network accuracy. However, optimization is needed to obtain similar inference times, given the higher model complexity.
Moreover, we aim to enhance the point cloud processing and the mapping procedure.
In particular, our current setup does not incorporate Simultaneous Localization and Mapping (SLAM) capabilities, therefore we are prone to drift over large scale environments. We are also thinking of using a monocular camera and learning-based, efficient depth estimation techniques for perceiving the environment, therefore avoiding the use of stereo cameras.
Additionally, we would like to develop autonomous exploration environment strategies. This process can still leverage multiple cost metrics employed in this work. By prioritizing areas with lower costs, we can autonomously guide the drone to directly explore areas that appears to be safer, making our drone entirely self-sufficient.
\section*{Acknowledgment}
We would like to thank Luca Morando and Pratyaksh P. Rao for their contributions and support to this project.
\bibliographystyle{IEEEtran}
\bibliography{biblio}







\end{document}